# Point Cloud Registration for Fusion between SPECT MPI and CTA Images

Ni Yao[1], Xiangyu Liu[1], Shaojie Tang[2,3], Danyang Sun[1], Chuang Han[1], Yanting Li[1], Jiaofen Nan[1], Chengyang Li[4], Fubao Zhu[1], Chen Zhao[5*], Zhihui Xu[6*], Weihua Zhou[7,8]

[1]School of Computer Science and Artificial Intelligence, Zhengzhou University of Light Industry, Zhengzhou 450001, Henan, China

[2]School of Automation, Xi'an University of Posts and Telecommunications, Xi'an, Shaanxi, 710121, China

[3]Xi'an Key Laboratory of Advanced Control and Intelligent Process, Xi'an, Shaanxi, 710121, China

[4]School of Information Management and Engineering, Shanghai University of Finance and Economics, China

[5]Department of Computer Science, Kennesaw State University Marietta, GA, USA

[6]Department of Cardiology, The First Affiliated Hospital of Nanjing Medical University, Nanjing, China

[7]Department of Applied Computing, Michigan Technological University, Houghton, MI, USA

[8]Center for Biocomputing and Digital Health, Institute of Computing and Cybersystems, and Health Research Institute, Michigan Technological University, Houghton, MI, USA

*Correspondence:

Chen Zhao, Ph.D.

Department of Computer Science

Kennesaw State University Marietta, GA, USA

Email: czhao4@kennesaw.edu

Zhihui Xu, M.D.

Department of Cardiology

The First Affiliated Hospital of Nanjing Medical University, Nanjing, China

Email: wx_xzh@njmu.edu.cn

# Abstract

Clinical fusion of Single Photon Emission Computed Tomography Myocardial Perfusion Imaging (SPECT MPI) and Computed Tomography Angiography (CTA) remains limited by cross-modality misregistration and reliance on manual landmarks, which can hinder accurate ischemia localization and lesion-level functional assessment. To address this, we propose a registration and fusion framework for SPECT MPI and CTA that integrates functional and structural information for comprehensive cardiac evaluation. The pipeline performs U-Net based segmentation on both modalities but adopts a different landmark strategy: on SPECT, only the left ventricle (LV) is extracted, and landmarks are automatically derived from characteristic LV anatomy; on CTA, both ventricles are extracted and their spatial relationship is used to automatically place landmarks at the interventricular septal junction. Scale-space consistency preprocessing and landmark-driven coarse registration mitigate initial misregistration. On this initialization, multiple fine-registration algorithms are applied to LV epicardial surface point clouds (ICP, SICP, CPD, CluReg, FFD, and BCPD++), and the resulting transformations are propagated to voxel-level resampling for high-precision SPECT–CTA fusion. In a retrospective cohort of 60 patients, the framework preserved sub-millimeter coronary detail from CTA while accurately overlaying quantitative SPECT perfusion; among the evaluated algorithms, BCPD++ achieved the highest accuracy with a mean point-cloud distance of 1.7 mm. By combining initialization, comparative fine registration, and voxel-level fusion, the approach provides a practical basis for myocardial ischemia localization and functional evaluation of coronary lesions, while remaining agnostic to any single fine-registration algorithm. Code: https://github.com/MIILab-MTU/SPECTMPIAndCTAFusion.

# 1. Introduction

In recent years, the incidence and mortality of cardiovascular diseases (CVDs) have continued to rise, making them a major global public health burden [1-5]. Factors such as population aging, lifestyle changes, and socioeconomic conditions have contributed to this trend [1 4]. Medical imaging plays a vital role in early detection, accurate diagnosis, and treatment monitoring, particularly with advances in imaging devices and algorithms [6]. In cardiovascular imaging, SPECT MPI and CTA are widely used modalities. SPECT enables quantitative evaluation of myocardial function but suffers from low resolution and motion artifacts [7-9]. CTA provides high-resolution coronary anatomy and is considered a gold standard for structural assessment, but it lacks functional information and has limitations in patients with contrast allergy or renal insufficiency [9 10]. Functional-anatomical fusion is clinically valuable, enabling simultaneous assessment of myocardial perfusion and coronary lesions in a single image. However, the fusion of SPECT and CTA faces multiple challenges: differences in imaging principles, cardiac motion, patient positioning, device parameters, and image noise all complicate registration [11 12]. Although landmark- and intensity-based methods have been explored, issues of computational complexity and robustness remain. We propose a point cloud-based coarse-to-fine registration and fusion approach for cardiac SPECT and CTA images. Coarse registration ensures that the cardiac orientation of SPECT and CTA images is correctly aligned. We introduce a cross-modality registration-and-fusion framework that standardizes initialization and enables comparative fine registration between SPECT and CTA. First, a U-Net is used to segment the LV epicardial region (LVER) in SPECT, and both the LVER and right ventricular epicardial region (RVER) in CTA. From CTA, we extract point clouds of the LV epicardial contours (LVECs) and the right ventricle, determine their intersection plane, and automatically select groove-like anatomical landmarks that delineate the interventricular junction. From SPECT, we also extract LVEC point clouds and, leveraging characteristic LVEC morphology, automatically select interventricular groove-like landmarks that correspond to the

CTA landmarks. Aligning these SPECT–CTA landmarks establishes a consistent cardiac orientation and provides a robust coarse alignment across modalities. Within this framework, multiple fine-registration algorithms are applied separately to the LV epicardial surface (LVES; i.e., all LVECs) point clouds—namely ICP[13], SICP[14], CPD[15], CluReg[16], FFD[17] and BCPD++[18]. The selected transformation from each algorithm is then propagated to the original 3D volumes for voxel-level resampling, yielding accurate SPECT–CTA registration and fusion. Notably, BCPD++ attains the highest registration accuracy within our framework, but the innovation lies in the asymmetric landmarking plus standardized pipeline that supports algorithm-agnostic, high-precision fusion rather than in any single fine-registration method.

Our main contributions are summarized as follows:

1. We propose a novel point cloud-based registration approach that combines anatomical landmark detection, scale-space consistency, and multi-stage refinement, achieving robust coarse-to-fine registration between SPECT and CTA images.

2. We develop automated algorithms for ventricular segmentation and landmark extraction across both modalities, enabling precise identification of the interventricular junction and the apical landmark.

3. Within the same standardized pipeline, we individually apply multiple fine-registration algorithms (ICP, SICP, CPD, CluReg, FFD, and BCPD++) to the LV epicardial surface point clouds. This plug-and-play design improves robustness and accuracy over single-method baselines, enables fair comparison across methods, and locates the best performer (BCPD++ in our study), while keeping the framework itself as the primary contribution.

The remainder of this article is organized as follows. Section 2 introduces the overall approach and methodological details. Section 3 presents the experimental setup, dataset, and results. Finally, Section 4 concludes this work.

# 2. Methods

## 2.1 Overall approach

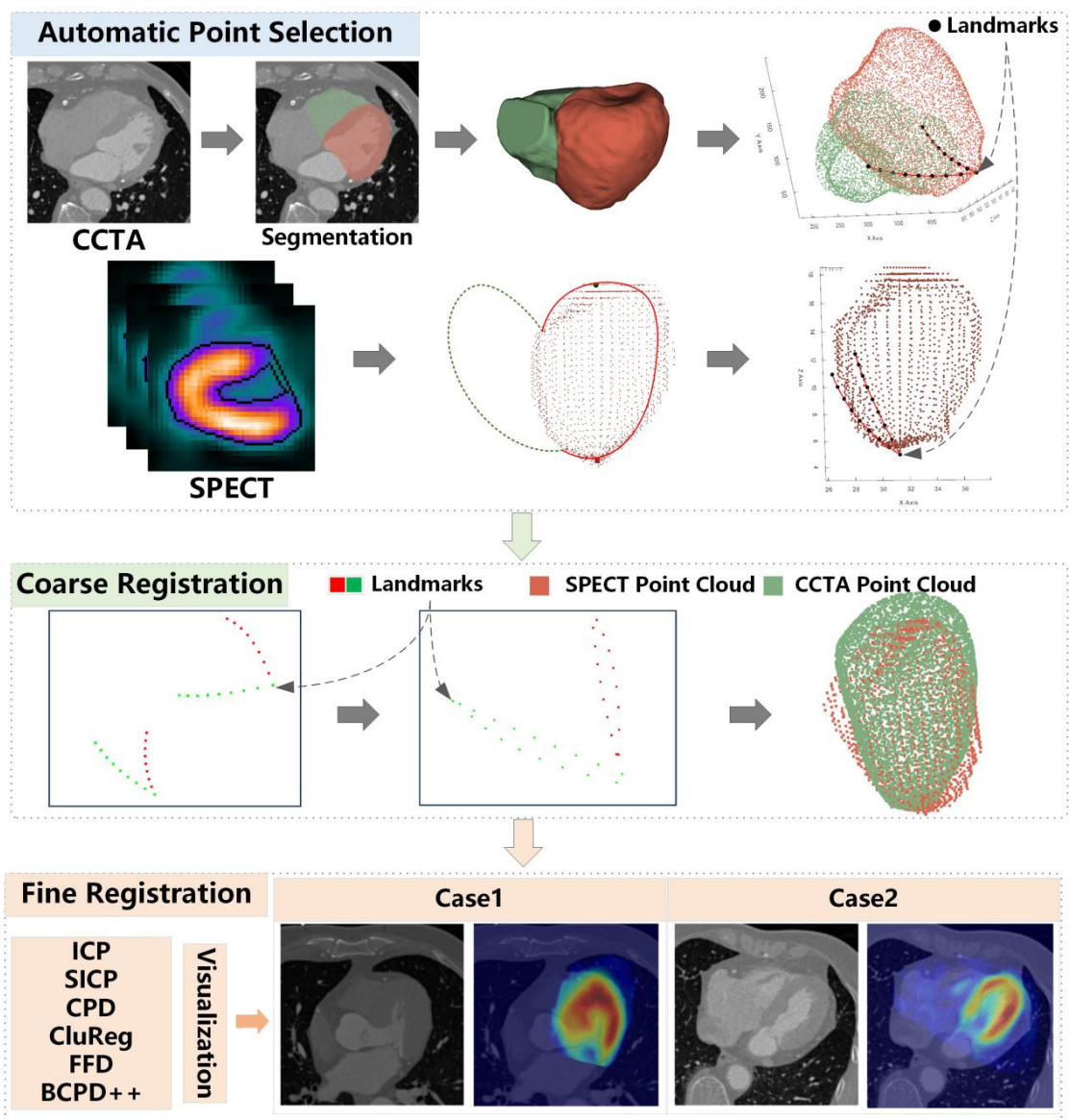

Figure 1. Overall process of CTA and SPECT image registration

We perform coarse-to-fine registration of CTA and SPECT images in the following three steps: **(1) Automatic point selection**. In this step, the LVERs of the SPECT image and the LVERs and RVERs of the CTA image were segmented, and the corresponding point clouds of the epicardium will be generated. Subsequently, the spatial relationship between the left and right ventricles in the CTA is utilized to identify landmarks where the left ventricular epicardial point cloud of the CTA intersects with the right ventricular epicardial. The corresponding landmarks where the right ventricle meets are determined by the anatomical characteristics of the left ventricular epicardial point cloud of the SPECT after point cloud preprocessing. **(2) Coarse registration**: Given the absence of substantial morphological disparities between the CTA and SPECT point clouds, direct registration would inevitably lead to registration errors. Therefore, it's necessary to register the two to their approximate positions. Subsequent to the acquisition of the landmarks obtained in step (1), coarse registration of the CTA and SPECT point clouds is implemented. **(3) Fine registration**: Algorithms such as ICP, SICP, CPD, CluReg, FFD and BCPD++ are employed to facilitate fine registration of the CTA and SPECT point clouds. Finally, the image registration results are obtained from the CTA and SPECT image registration. The overall workflow is illustrated in Figure 1.

**2.2 Automatic Selector**

2.2.1 Ventricular segmentation

We segmented the ventricles in both CTA and SPECT images to generate high-quality data for subsequent point cloud extraction and analysis. For CTA data, we employed a U-Net model to segment the left and right ventricles. After training, the U-Net was applied to extract the outer membrane of the left ventricle as point cloud data for further analysis. The U-Net architecture [19] is illustrated in Figure 4. The segmentation results were rendered as surface point clouds using VTK, as shown in Figure 2.

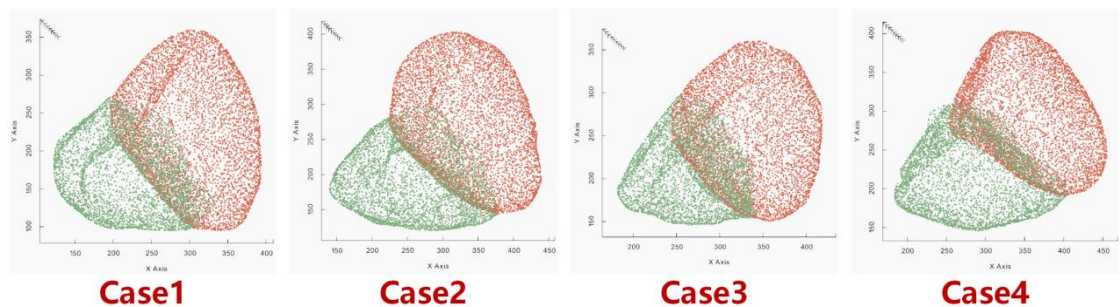

Figure 2. Visualization of point clouds formed by the right and left ventricles after CTA image segmentation, shown for four cases. The green point cloud is generated from the segmented right ventricle and the red point cloud is generated from the segmented left ventricle.

For SPECT data processing, an algorithm was designed to precisely locate anatomical landmarks of the left ventricle in SPECT images, including the center point, apex, and base. The original short-axis (SAX) images were resampled into a long-axis (LAX) perspective and rotated at 9° intervals to generate 20 slice planes. All sampled planes were then mapped to the polar coordinate system via coordinate transformation, thereby reducing geometric distortion and improving the accuracy of subsequent processing. On this basis, an automated segmentation procedure was implemented using a pre-trained U-Net model [20] for extraction of the left ventricle. Finally, the segmented outer membrane of the left ventricle was mapped back to the original Cartesian coordinate system through inverse transformation, ensuring strict consistency between the generated point cloud and the original SPECT image. The resulting point cloud of the segmented left epicardium is shown in Figure 3.We visualized partial coronary and transverse views of the resampling process using the segmented ventricular outer membrane point cloud, as shown in Figure 5.The sampling process was divided into 20 iterations, and we visualized the 1st, 2nd, 5th, 7th, and 20th iterations.

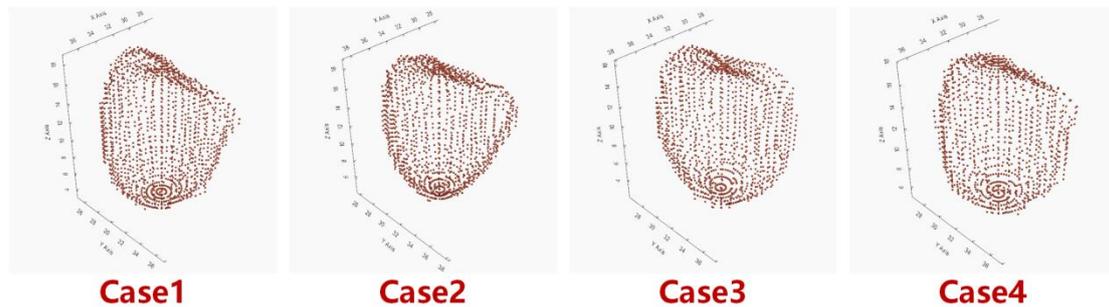

Figure 3. Visualization of point clouds formed by the left ventricle after SPECT image segmentation, shown for four cases. The red point cloud is generated by the segmented left ventricle.

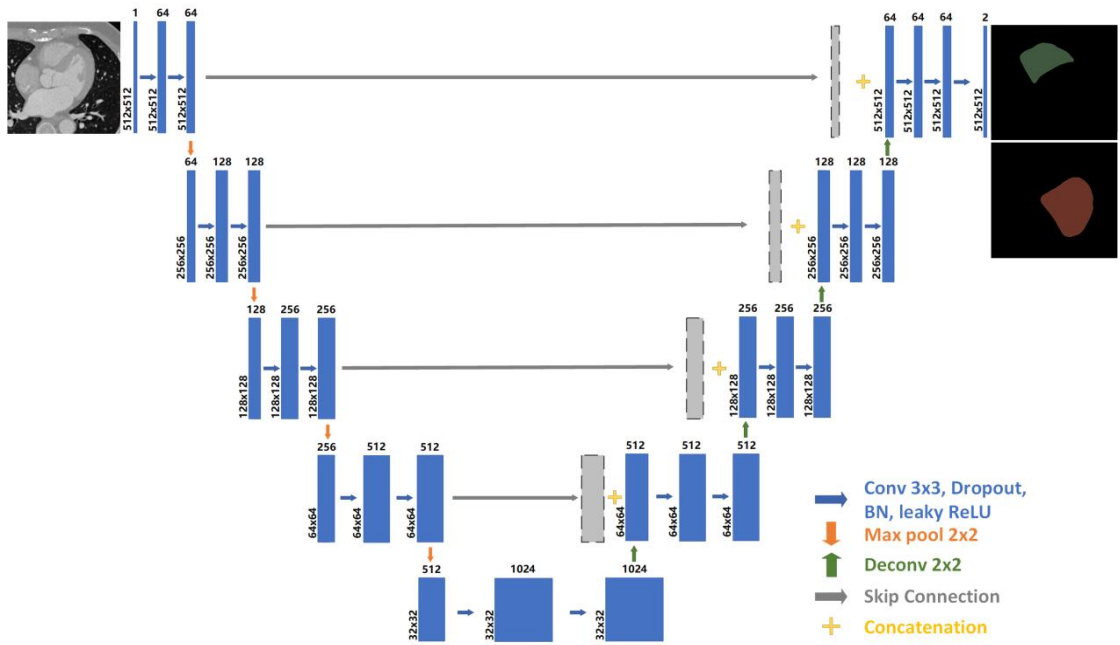

Figure 4. U-net segmentation network

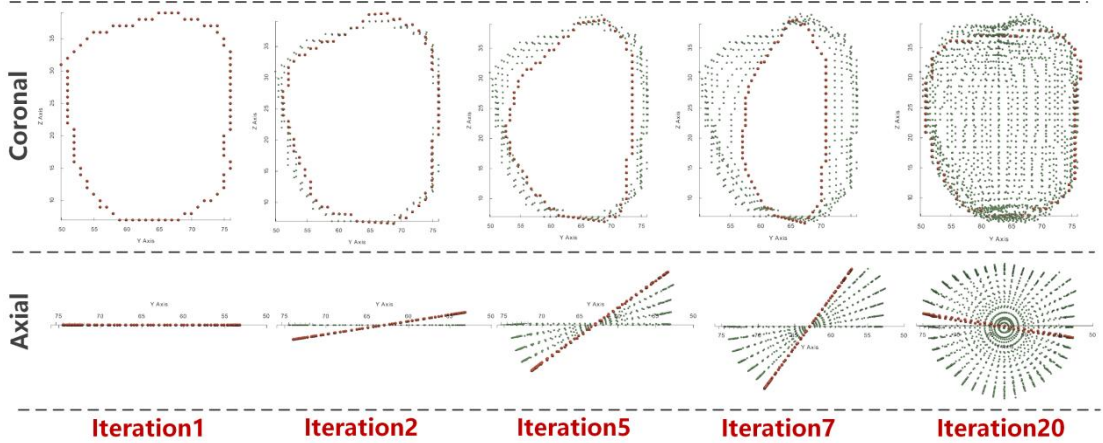

Figure 5. Sampling process of the SPECT left epicardial point cloud. The red points represent the points added during this step, while the green points represent the points that have already been sampled.

2.2.2 Automatic labeling of CTA left epicardial landmarks

In this step, we propose an automatic calculation method for anatomical junction planes based on principal component analysis (PCA) of contact areas, which is used to robustly estimate the anatomical separation interface between the left and right epicardial point clouds. A dual k-d tree[21] nearest neighbor query is applied to symmetrically extract the contact point sets of the left and right ventricles within a predefined Euclidean distance threshold, thereby constructing a local point cloud that reflects the true contact region. If there are insufficient contact points, the algorithm degrades to performing PCA on the entire cardiac cavity point cloud to obtain a global approximate plane. Next, the zero-mean covariance matrix is calculated for the contact point set, and the eigenvector corresponding to the smallest eigenvalue is used as the normal vector of the interface plane, with the centroid of the point set designated as the plane center. This ensures that the plane spatially traverses the contact region and is orthogonal to its maximum extension direction.

After obtaining the anatomical interface planes of the left and right ventricle chambers, the left ventricular epicardial point cloud is orthogonally projected onto the plane to construct a two-dimensional parametric representation. Subsequently, two-dimensional principal component analysis is performed within the projection domain to extract the long axis and short axis directions, thereby characterizing the morphological axes of the ventricle chambers within the junction plane. The endpoints of the long axis are determined by the extrema of the projection points along this principal axis, while the apex point is determined by the curvature of the local point cloud boundary at the endpoints. Based on the midpoint and direction of the long axis, a short axis plane perpendicular to the junction plane is constructed; The intersection line between this plane and the original intersection plane defines the potential intersection line direction. To precisely locate the anatomical intersection, the algorithm performs a bidirectional contraction search along the intersection line direction: first, candidate points within the short axis plane neighborhood that satisfy the intersection plane distance tolerance are screened; then, the search gradually moves inward along the intersection line direction until the candidate points satisfy the proximity constraint with the left ventricle point cloud, thereby generating a pair of adjusted intersection points. Finally, symmetrical Bézier curves are generated from the apex point to the candidate points, and 9 points are uniformly sampled along each curve as the special intersection points between the left and right ventricle point clouds in the CTA. This process is visualized in Figure 6.

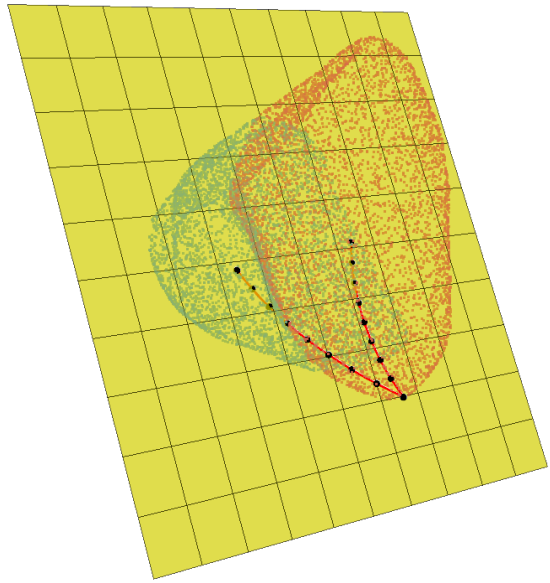

Figure 6. Sampling process of the CTA landmarks. The color red indicates the CTA left epicardial point cloud, while green points represent the CTA right epicardial point cloud. The yellow plane signifies the anatomical junction plane between the left and right ventricle chambers. The blue color denotes the generated Bézier curve, and black represents the uniformly sampled landmarks.

2.2.3 Automatic labeling of SPECT left epicardial landmarks

(1) We introduce a center point estimation step based on Randomized Consistency (RANSAC) prior to identifying the apex and base. Leveraging the structural organization of the SPECT left epicardial point cloud, which is uniformly sampled every 20° around the circumference, we first orthogonally project the cardiac cavity point cloud onto the XY plane. A unique coordinate deduplication strategy is then applied to remove duplicate sampling points, thereby reducing subsequent fitting errors. We subsequently use RANSAC to perform iterative circle fitting, prioritizing models that contain the largest number of internal points and enforcing a zero count of circumferential points to ensure that the fitted circle is supported entirely by internal points rather than edge noise. Finally, the apical and basal points are identified based on the obtained center coordinates $(x_c,\ y_c)$.

(2) We propose an automatic extraction strategy for the SPECT left epicardium-approximate right ventricle junction plane based on long-axis rotation sampling and maximization of coverage density. First, we construct the long axis of the ventricle by connecting the left apex and basal centers, and generate an initial basis vector orthogonal to this axis as the rotation center. By rotating this basis vector around the long axis in 5° increments from 0° to 360°, we obtain a series of candidate plane normal vectors. The plane center is then set at 60% of the maximum projection radius from the long-axis center, and the number of points covered by each plane is calculated. The plane with the maximum coverage and no boundary points on its circumference is selected as the optimal junction plane.

(3) The short axis plane extraction method is the same as that of CTA. Finally, symmetrical Bézier curves are generated from the apex point to the candidate points, and 18 points are uniformly sampled as landmarks approximating the left and right ventricle point clouds in SPECT. For ease of understanding, we have illustrated this process in Figure 7.

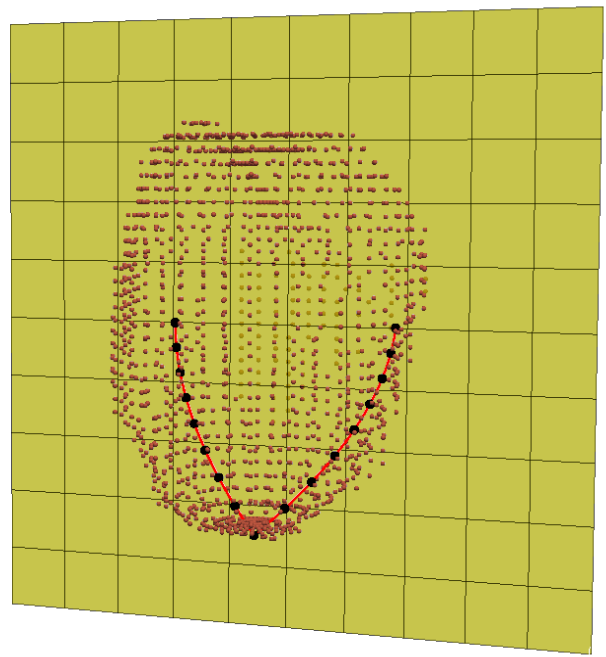

Figure 7. Sampling process of the SPECT landmarks. The color red indicates the SPECT left epicardial point cloud, The color yellow indicates the approximate intersection plane of the SPECT left and right ventricle chambers, The color blue indicates the generated Bézier curve, and black indicates the uniformly sampled landmarks.

**2.3 Coarse registration**

Due to the lack of significant morphological landmarks in the left ventricle (LV) itself, conventional registration of LV EC point clouds from SPECT and CTA is prone to misalignment. As shown in Figure 8, the contour overlap between the two modalities is low, resulting in significant deviations in the registration results. To enable comparison and registration of SPECT myocardial perfusion point clouds and CTA anatomical point clouds within the same three-dimensional coordinate system, this study introduces a 'scale-space consistency' preprocessing technique prior to formal registration. This technique consists of two steps: scale normalization and initial alignment based on anatomical landmarks, thereby avoiding the deviations caused by direct registration.

First, considering the differences between SPECT and CTA in imaging physics, reconstruction algorithms, and voxel size, the two-modality point clouds often exhibit overall scale deviations. To address this, we calculate the zero-mean covariance matrices for both point clouds and extract the eigenvalues along the three principal axes. By comparing the ratios of variances along the corresponding principal axes, we derive an isotropic average scale factor to uniformly scale the SPECT point cloud up or down, ensuring it matches the CTA in terms of volume and extent.

Subsequently, to obtain a reliable initial rigid transformation, we utilize a pre-sampled set of anatomical landmarks from CTA and SPECT. These landmarks have been synchronously scaled up or down during the scale normalization stage, ensuring that their relative positions remain unchanged. Using the Umeyama algorithm[22], the rotation matrix and translation vector can be analytically solved by minimizing the sum of squared Euclidean distances between corresponding point pairs, thereby completing the first rigid alignment.

Finally, the obtained initial transformation matrix is applied to the entire SPECT point cloud in one go, achieving a seamless transition from 'scale-space consistency' preprocessing to 'anatomical landmark-driven' coarse registration.

The point cloud comparison before and after coarse registration based on landmarks is shown in Figure 8. In the figure, the SPECT point cloud is displayed in red, and the CTA point cloud is displayed in green. In addition, based on coarse registration, four algorithms, ICP, SICP, CPD, and BCPD++, were applied, and the fine registration results are shown in Figure 9. The colors of the points in the figure have the same meaning as in Figure 8.

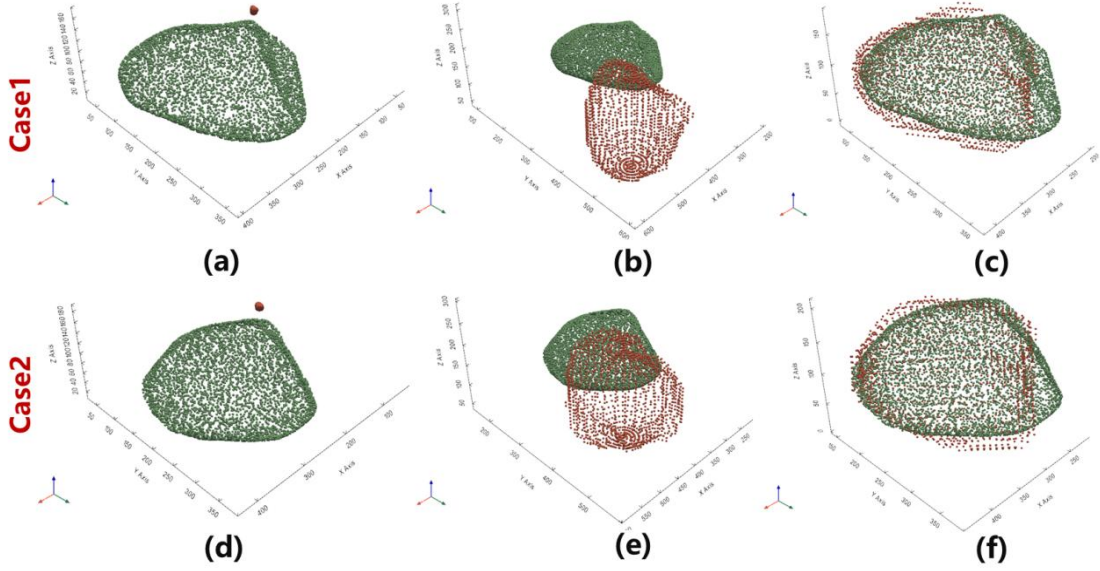

Figure 8. Visualization of the coarse registration process for two point clouds, where red represents the SPECT left epicardial point cloud and green represents the CTA left epicardial point cloud. (a) shows the two point clouds in their initial unregistered state, (b) shows the SPECT point cloud after magnification, and (c) shows the result after applying the special point registration extracted.

**2.4 Fine registration**

In the fine-tuning stage of myocardial point cloud registration, this study sequentially applied the Iterative Closest Point (ICP) algorithm, the Sparse Iterative Closest Point (SICP) algorithm, the CPD Rigid algorithm, CPD Affine, and BCPD to achieve high-precision alignment of SPECT MPI and CTA point clouds.

The ICP (Iterative Closest Point) algorithm [13] is a classic point cloud registration method that aligns the source point cloud and target point cloud through an iterative process. The basic assumption of this method is that the source point cloud and target point cloud can be aligned under rigid transformations involving rotation and translation. In each iteration, the ICP algorithm identifies the nearest neighbor point pairs between the source and target point clouds, calculates the optimal rotation matrix and displacement vector, and minimizes the distance between them. The minimization objective function of the ICP algorithm is shown in Formula (1).

$$E(\mathbf{R}, \mathbf{t}) = \sum_{i=1}^{N} \| \mathbf{R}\mathbf{p}_i + \mathbf{t} - \mathbf{q}_i \|^2 \tag{1}$$

Where $\mathbf{R}$ is the rotation matrix, $N$ is the number of target point clouds, $\mathbf{t}$ is the translation vector, $\mathbf{p}_i$ and $\mathbf{q}_i$ are the $i$-th point in the source and target point clouds respectively. The algorithm solves the rotation matrix and translation vectors by minimizing this error function and finally achieves the point cloud registration.

The SICP (Scaled Iterative Closest Point) algorithm[14] is an extension of the ICP algorithm, introducing a scale factor to handle scale differences between the source point cloud and the target point cloud. Unlike the standard ICP algorithm, which assumes that point clouds are consistent in scale, the SICP algorithm introduces a scale transformation matrix to enable the algorithm to adapt to different scales. The objective function of the SICP algorithm is shown in Formula (2).

$$E(\mathbf{R}, \mathbf{t}, \mathbf{s}) = \sum_{i=1}^{N} \| \mathbf{sRp}_i + \mathbf{t} - \mathbf{q}_i \|^2 \qquad (2)$$

Where $\mathbf{s}$ is the scale factor, $\mathbf{R}$ is the rotation matrix, $N$ is the number of target point clouds, $\mathbf{t}$ is the translation vector, and $\mathbf{p}_i$ and $\mathbf{q}_i$ are the $i$-th points in the source and target point clouds, respectively. By minimizing the error function, the SICP algorithm is able to estimate the rotation, translation and scale transformation parameters to achieve the alignment of the point cloud.

The CPD (Coherent Point Drift) algorithm [15] is a point cloud registration method based on a probabilistic model, which achieves registration by maximizing the alignment probability between the source point cloud and the target point cloud. CPD Rigid is a variant of the CPD algorithm, focusing on point cloud registration for rigid transformations (rotation and translation). Each point in the point cloud is modelled as a Gaussian distribution, and the source point cloud is aligned with the target point cloud using a rotation matrix and a translation vector. The transformation formula and maximization objective function for CPD Rigid are given by Equations (3) and (4), respectively.

$$\mathbf{q}_i = \mathbf{Rp}_i + \mathbf{t}, \forall i = 1,2,\ldots,N \qquad (3)$$

$$\mathcal{L}(\mathbf{R}, \mathbf{t}) = \prod_{i=1}^{N} \exp\left(-\frac{\|\mathbf{q}_i - (\mathbf{Rp}_i + \mathbf{t})\|^2}{2\sigma^2}\right) \qquad (4)$$

Where, $\mathbf{q}_i$ is the $i$-th point in the target point cloud, $\mathbf{p}_i$ is the $i$-th point in the source point cloud, $\mathbf{R}$ is the rotation matrix, $\mathbf{t}$ is the translation vector, and $\sigma^2$ is the noise variance, $N$ is the number of target point clouds.

CPD Affine is an extension of the CPD algorithm that allows affine transformations between point clouds. Unlike rigid transformations, affine transformations also include scaling and shearing operations, enabling CPD Affine to handle more complex point cloud alignment problems. The transformation formula and maximization objective function for CPD Affine are given by Equations (5) and (6), respectively.

$$\mathbf{q}_i = \mathbf{A}\mathbf{p}_i + \mathbf{b}, \quad \forall i = 1,2,\dots,N \tag{5}$$

$$\mathcal{L}(\mathbf{A}, \mathbf{b}) = \prod_{i=1}^{N} \exp\left(-\frac{\|\mathbf{q}_i - (\mathbf{A}\mathbf{p}_i + \mathbf{b})\|^2}{2\sigma^2}\right) \tag{6}$$

Where $\mathbf{q}_i$ is the $i$-th point in the target point cloud, $\mathbf{p}_i$ is the $i$-th point in the source point cloud, $\mathbf{A}$ is the affine transformation matrix, and $\mathbf{b}$ is the translation vector, $N$ is the number of target point clouds.

The BCPD++ (Bayesian Coherent Point Drift++) algorithm[18] is a point cloud registration method based on Bayesian inference that can handle rigid, affine, and non-rigid transformations. BCPD++ incorporates a Bayesian framework to simultaneously account for noise and local deformations in point cloud data during the process of maximizing the posterior probability, thereby enhancing the robustness of point cloud registration. The non-rigid transformation component is modelled using a Gaussian process model, enabling BCPD++ to handle more complex deformations. The transformation formula and maximization objective function for BCPD++ are given by Equations (7) and (8), respectively.

$$\mathbf{q}_i = \mathbf{R}\mathbf{p}_i + \mathbf{t} + \delta\mathbf{p}_i, \quad \forall i = 1,2,\dots,N \tag{7}$$

$$\mathcal{L}(\mathbf{R}, \mathbf{t}, \delta\mathbf{p}_i) = \prod_{i=1}^{N} \exp\left(-\frac{\|\mathbf{q}_i - (\mathbf{R}\mathbf{p}_i + \mathbf{t} + \delta\mathbf{p}_i)\|^2}{2\sigma^2}\right) \tag{8}$$

wherein $\mathbf{q}_i$ is the $i$-th point in the target point cloud, $\mathbf{p}_i$ is the $i$-th point in the source point cloud, $\mathbf{R}$ is a rotation matrix, $\mathbf{t}$ is a translation vector, and $\delta\mathbf{p}_i$ is a non-rigid transform portion, $N$ is the number of target point clouds.

The CluReg (Cluster-based Registration) algorithm [16] is a non-rigid point cloud registration method that partitions the point set into clusters and estimates local transformations in each region. By combining local rigid or affine models with smooth weighting, CluReg achieves global non-rigid alignment while maintaining local adaptability. The transformation for a point $\mathbf{p}_i$ is defined as Equations (9).

$$\mathbf{q}_i = \sum_{k=1}^{K} w_k(\mathbf{p}_i)(\mathbf{R}_k \mathbf{p}_i + \mathbf{t}_k), \quad i = 1, 2, \ldots, N \tag{9}$$

where $\mathbf{q}_i$ is the transformed position of the source point $\mathbf{p}_i$, $\mathbf{R}_k$ and $\mathbf{t}_k$ denote the rotation and translation of the k-th cluster, and $w_k(\mathbf{p}_i)$ is the normalized weight function ($\sum_k w_k(\mathbf{p}_i) = 1$) controlling the influence of each cluster, shown as Equations (10).

The optimization objective is formulated by minimizing the weighted alignment error between the transformed source points and the corresponding target clusters.

$$\mathscr{L}(\mathbf{R}_k, \mathbf{t}_k) = \sum_{i=1}^{N} \sum_{k=1}^{K} w_k(\mathbf{p}_i) \|\mathbf{q}_i - (\mathbf{R}_k \mathbf{p}_i + \mathbf{t}_k)\|^2 \tag{10}$$

This formulation ensures that local transformations are smoothly blended, avoiding discontinuities at cluster boundaries.

The FFD (Free-Form Deformation) [17] algorithm is a non-rigid registration approach that employs a lattice of control points and B-spline interpolation to model smooth deformations. The transformation of a point $v = (x, y, z)$ with respect to the control grid is expressed as Equations (11).

$$T(x, y, z) = v + \sum_{i=0}^{l} \sum_{j=0}^{m} \sum_{k=0}^{n} B_i(u) B_j(v) B_k(w) \Delta P_{i,j,k} \tag{11}$$

where $(u, v, w)$ are the normalized local coordinates of x, $\Delta P_{i,j,k}$ is the displacement of the control point at grid location $(i, j, k)$, and $B_i(\cdot)$ denotes the cubic B-spline basis function.

The optimization seeks to minimize the energy functional composed of two terms: the alignment error between the transformed source and the target point cloud, and a regularization term penalizing excessive control point displacements to ensure smoothness, shown as Equations (12).

$$\mathcal{E}(\Delta P) = \sum_{i=1}^{N} \| q_i - T(p_i) \|^2 + \lambda \, \|\nabla^2 \Delta P\|^2 \qquad (12)$$

where $\lambda$ is the regularisation weight. This ensures the deformation field is both accurate and anatomically plausible. $q_i$ is the transformed position of the source point $p_i$.

Furthermore, the transformation matrix generated during the registration process includes a rotation matrix, a translation vector, and a non-rigid transformation component, where the non-rigid transformation component is represented by an offset matrix. In addition, based on coarse registration, four algorithms, ICP, SICP, CPD, CluReg, FFD and BCPD++, were applied, and the fine registration results are shown in Figure 9. The colors of the points in the figure have the same meaning as in Figure 8.

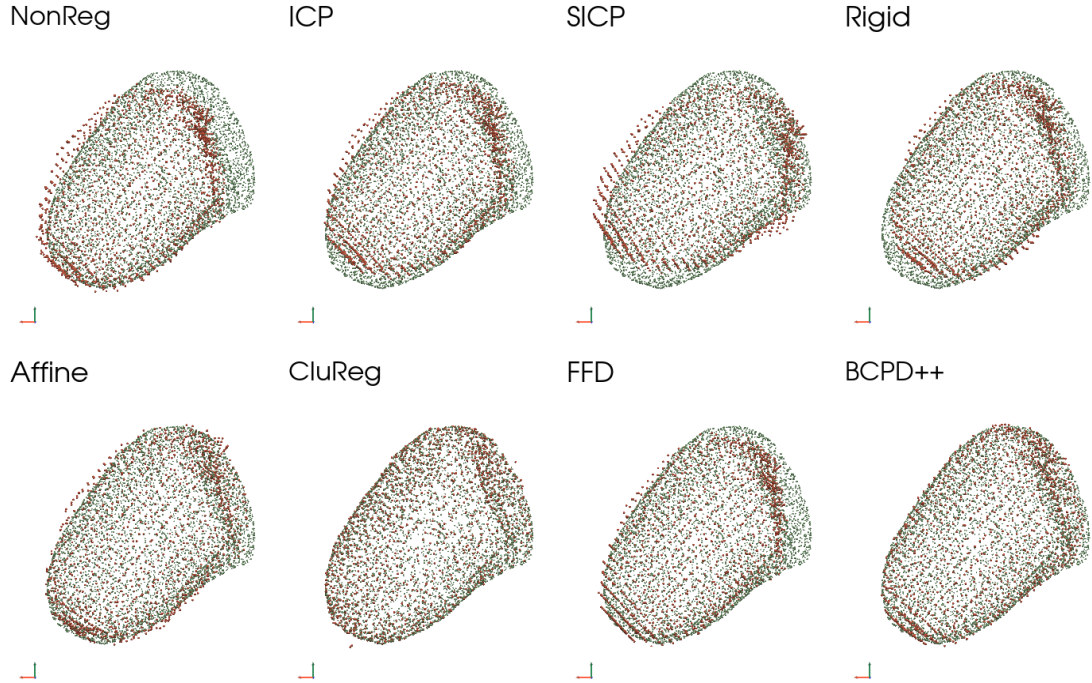

Figure 9. A visualization of the results of point cloud fine registration, where NonReg is the result of coarse registration.

**2.5 Image registration**

After point cloud registration, the next step is image-level registration. Using the transformation matrices obtained from point cloud registration (including rigid transformations, affine transformations, and non-rigid transformations), we apply

these transformations to register SPECT images and CTA images. The specific steps are as follows:

First, generate the grid coordinates of the target image (CTA) and use the obtained transformation matrix (including the rotation matrix and translation vector) to perform an inverse transformation on the grid, thereby determining the transformation relationship between the source image (SPECT image) and the target image (CTA image). For the BCPD++ method, we use the nearest-neighbor interpolation (NNI) method to interpolate the offset matrix and generate the deformation field. To ensure high-quality image registration, cubic spline interpolation was used to resample the SPECT image, enabling precise alignment between the SPECT image and the CTA image.

## 3. Experiments and Results

### 3.1 Data and Experimental Setup

All data used in this experiment were obtained from the Jiangsu Provincial People's Hospital. We used a U-Net model to segment the left and right ventricles in CTA images. A total of 16,132 CTA images from 66 patients were divided into training, validation, and test sets. Among these, 13,231 slices from 54 patients were used for training and validation, while the remaining 2,901 slices from 12 patients were used as an independent test set for model evaluation. All slices were uniformly resized to 512×512 pixels to ensure data consistency and to meet the model input requirements. In the SPECT segmentation task, pre-trained weights were used to segment 60 cases of SPECT data. For point cloud registration and image registration tasks, we used 60 patients who had both SPECT and CTA data. Before performing the CTA segmentation task, all images were subjected to Max-Min normalization. The training epoch was set to 100, the optimizer was Adam, and the initial learning rate was 0.01. Binary cross entropy (BCE) loss was employed in the segmentation process, as shown in Formula (13).

$$\mathcal{L}_{\text{BCE}} = -\frac{1}{N}\sum_{i=1}^{N}(y_i \log(\hat{y}_i) + (1 - y_i) \log(1 - \hat{y}_i)) \tag{13}$$

Where N is the number of pixels in the $N$ image. $y_i$ is the true label of the $i$-th pixel. $\hat{y}_i$ is the prediction of the $i$-th pixel.

All experiments were conducted on a unified hardware platform to ensure comparability and reproducibility of the results. The experimental environment consisted of an AMD Ryzen 9 9950 processor, 128 GB of memory, and an NVIDIA RTX 5090D GPU with 32 GB of VRAM.

**3.2 Evaluation Metrics**

We employed the Dice similarity coefficient (DSC) to evaluate the segmentation results, as shown in Formula (14).

$$\text{Dice} = \frac{2\sum_{i=1}^{N} p_i \, g_i}{\sum_{i=1}^{N} p_i + \sum_{i=1}^{N} g_i} \tag{14}$$

Where $p_i$ is the probability that the $i$-th pixel is predicted to be foreground, $g_i$ is the probability that the $i$-th pixel is truly labeled as foreground, and $N$ is the total number of pixels in the image.

To ensure the effectiveness of point cloud registration and image registration, we used mean point cloud distance error (MPE), apex displacement angles error (AE), mean interventricular groove distance error (MGE) and interventricular groove plane center error (GCE) to evaluate the registration accuracy. The point cloud average distance refers to the average distance between the CTA left epicardial point cloud and the registered SPECT left epicardial point cloud, as shown in Formula (15). The point cloud average distance quantifies the overall geometric deviation between the epicardial point clouds of the left ventricle in CTA and registered SPECT images. It reflects the global spatial consistency of the two modalities, where smaller values indicate higher registration accuracy.

$$D_{\text{avg}} = \frac{1}{N} \sum_{i=1}^{N} \| \mathbf{p}_i - \mathbf{q}_i \|_2 \tag{15}$$

where $N$ is the number of points in the point cloud, $\mathbf{p}_i$ denotes the coordinates of the $i$-th point in the CTA epicardial point cloud, and $\mathbf{q}_i$ denotes the coordinates of the corresponding point in the aligned SPECT left epicardial point cloud.

We manually marked the left ventricular apex points on the CTA images and

SPECT images. After registering the SPECT apex points with the point cloud, we calculated the angular displacement between the SPECT apex points and the CTA apex points. The angular displacement is calculated as shown in Formula (16). The angular displacement evaluates the consistency of cardiac orientation by calculating the angle between the apical vectors of CTA and SPECT. It is sensitive to rotational or directional mismatches, even when global distance errors are small. Accurate alignment of cardiac orientation is critical for myocardial segmentation, perfusion defect localization, and precise mapping of coronary territories.

$$\theta_{apex} = \arccos\left(\frac{\mathbf{v}_{CTA} \cdot \mathbf{v}_{SPECT}}{\|\mathbf{v}_{CTA}\|_2 \|\mathbf{v}_{SPECT}\|_2}\right) \quad (16)$$

Where $\mathbf{v}_{CTA} = p_{apex,CTA} - p_{center,CTA}$ and $\mathbf{v}_{SPECT} = p_{apex,SPECT} - p_{center,CTA}$, $p_{apex,CTA}$ is the apex of CTA, $p_{center,CTA}$ is the center point of CTA point cloud, and $p_{apex,SPECT}$ is the apex of SPECT.

We also manually labeled the interventricular grooves points in both the CTA and SPECT images. After registering the SPECT interventricular groove points with the point cloud, we calculated the average distance between the SPECT interventricular groove points and the CTA interventricular groove points. The calculation method is the same as in Formula (15). Using the interventricular groove as an anatomical landmark, this metric measures the degree of alignment between CTA and SPECT at the ventricular boundary. As the interventricular groove is a clinically significant divider between the left and right ventricles, this metric captures the accuracy of registration in a region of high diagnostic relevance, complementing global measures.

### 3.3 Results and Analysis

The U-Net achieved a DSC of 0.9531 for the left-ventricle segmentation and 0.8460 for the right-ventricle segmentation on the test set.

For the point cloud registration task, we compared the mean point cloud distance error (MPE), apex displacement angles error (AE), mean interventricular groove distance error (MGE) and interventricular groove plane center error (GCE) of five methods: ICP, SICP, CPD_Rigid, CPD_Affine, and BCPD++. The specific results are shown in Table 1.

Table 1 Results of coarse registration followed by fine registration

|         | ICP  | SICP | CPD_Rigid | CPD_Affine | CluReg | FFD | BCPD++ |
|---------|------|------|-----------|------------|--------|-----|--------|
| MPE(mm) | 3.8  | 3.6  | 3.2       | 2.2        | 2.5    | 2.5 | **1.7** |
| AE(°)   | 12.0 | 24.8 | 12.3      | 11.3       | 9.7    | 7.9 | **6.7** |
| MGE(mm) | 9.6  | 17.9 | 13.0      | 11.8       | 9.5    | 8.4 | **8.0** |
| GCE(mm) | 8.9  | 15.3 | 14.6      | 11.9       | 8.3    | **6.5** | 6.6 |

Based on the proposed coarse-to-fine registration method, we systematically evaluated the performance of multiple point cloud registration algorithms across four quantitative indicators: MPE, AE, MGE, and GCE. Among all tested methods, BCPD++ consistently achieved the best overall performance across the four metrics. Specifically, using BCPD++ reduced the MPE to 1.7 mm, outperforming ICP 3.8 mm and all other rigid or affine baselines. The AE was also minimized to 6.7°, representing a remarkable reduction of 5.3° compared with ICP and far lower than SICP 24.8° or CPD_Affine 11.3°. For interventricular groove alignment, BCPD++ decreased MGE to 8.0 mm and the GCE to 6.6 mm. Although FFD achieved a lower GCE 6.5 mm, its errors in MPE 2.5 mm and MGE 8.4 mm revealed less stability and overall consistency. Similarly, CluReg achieved competitive accuracy in GCE 8.3 mm, but its AE remained larger 9.7°. In contrast, BCPD++ provided a well-balanced and robust improvement across all anatomical landmarks and orientation metrics, reducing both local and global misalignment simultaneously.

Based on the volume and voxel size parameters of the SPECT and CTA images, we constructed spatial reference objects for the SPECT and CTA images and combined the results of point cloud registration to transform and register the original 3D SPECT images. The image registration results are illustrated in three anatomical planes: sagittal (Figure 10), transverse (Figure 11), and coronal (Figure 12).

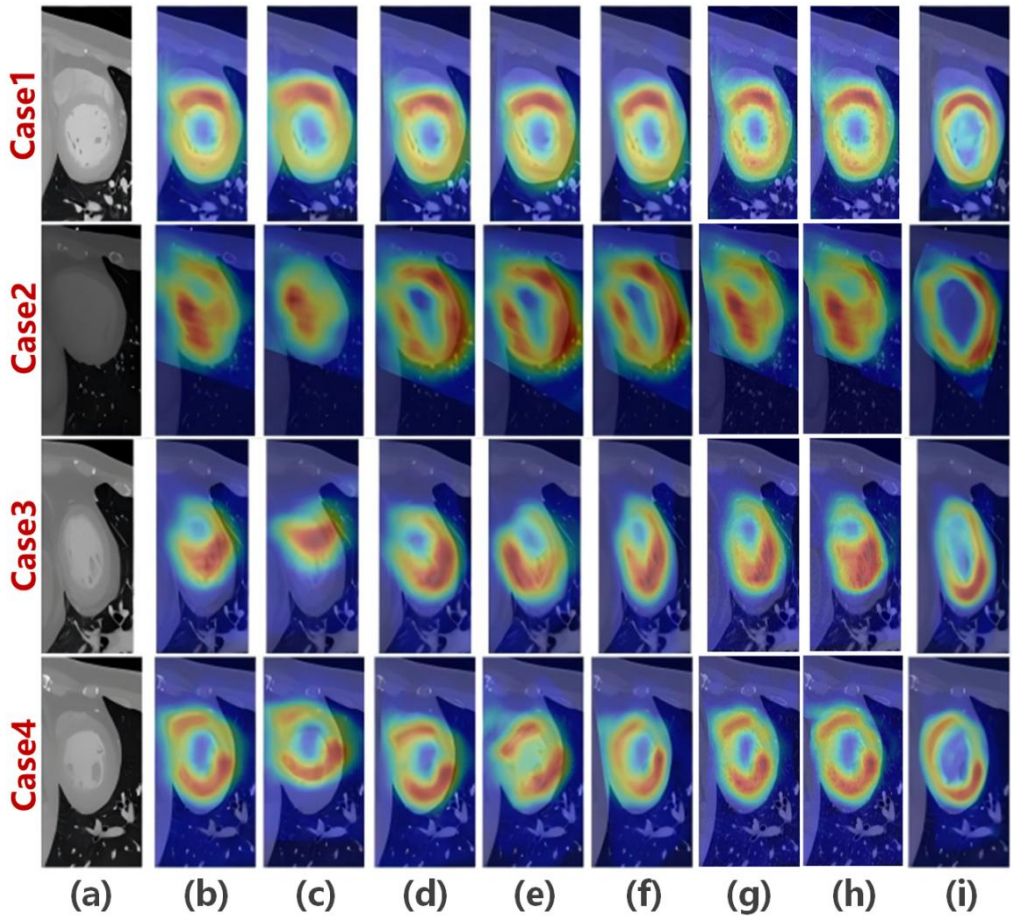

Figure 10. Visualization of the sagittal view of the registration results of four CTA and SPECT images, where (a) is the CTA, (b) is the rough registration result, (c) is the SICP registration result, (d) is the ICP registration result, (e) is the CPD Rigid registration result, (f) is the CPD Affine registration result, (g) is the CluReg registration result, (h) is the FFD registration result, and (i) is the BCPD++ registration result.

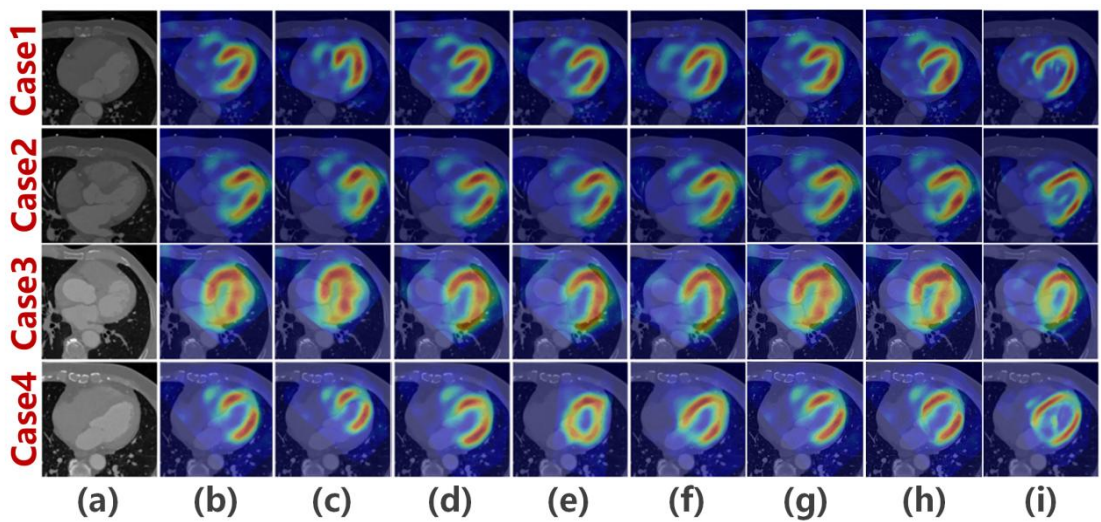

Figure 11. Visualization of cross-sectional views of four CTA and SPECT image registration results, (a-f) consistent with Figure 10.

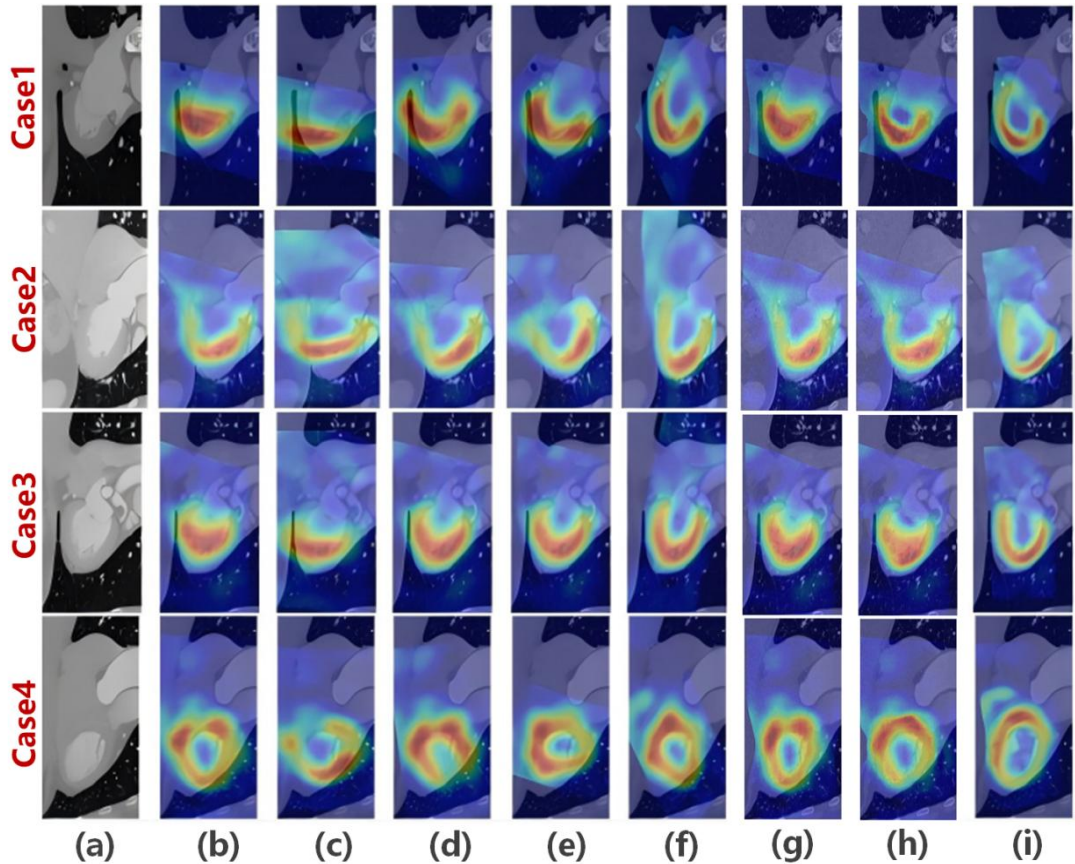

Figure 12. Visualization of the results of CTA and SPECT image registration in the coronal view for four cases, (a-f) consistent with Figure 10.

Figures 10-12 present the visualization of CTA-SPECT registration results across sagittal (Figure 10), cross-sectional (Figure 11), and coronal (Figure 12) planes for four representative cases. Panel (a) shows the CTA reference, and panels (b-i) display the outcomes of different registration strategies. Overall, rigid methods such as ICP, SICP and CPD_rigid produced suboptimal alignment, with noticeable residual offsets between perfusion hotspots and myocardial boundaries. In contrast, CPD_Affine achieved relatively good performance, effectively capturing global scale and orientation differences and yielding improved overlap compared with purely rigid methods. This demonstrates its utility in scenarios where affine transformations dominate the inter-modality discrepancy.

Nevertheless, the most significant improvements were obtained with non-rigid approaches. CluReg and FFD reduced local mismatches to varying degrees, though their performance was somewhat case-dependent. BCPD++ consistently delivered the most accurate and stable alignment across all three views, ensuring that the SPECT perfusion signals were well-registered to the CTA anatomical structures. This superior performance highlights BCPD++'s ability to model complex local deformations while preserving global anatomical integrity, thereby offering the most reliable fusion.

In order to demonstrate the necessity of coarse registration before fine registration, we directly performed fine registration on CTA left endocardium point cloud and SPECT left endocardium point cloud, and the results are shown in Table 2.

Table 2 Results of fine registration without coarse registration

|         | ICP   | SICP  | CPD_Rigid | CPD_Affine | CluReg | FFD   | BCPD++ |
|---------|-------|-------|-----------|------------|--------|-------|--------|
| MPE(mm) | 4.6   | 4.5   | 3.3       | 2.5        | 1.9    | 39.2  | **1.6**|
| AE(°)   | 116.8 | 98.5  | 56.7      | 83.4       | 78.9   | 114.2 | **51.1**|
| MGE(mm) | 39.1  | 195.2 | 31.5      | 33.4       | **31.0** | 68.6 | 38.6  |
| GCE(mm) | 51.3  | 207.3 | **39.7**  | 46.9       | 47.5   | 82.8  | 48.7   |

Table 2 presents the results of fine registration without the aid of coarse registration. In this setting, all algorithms exhibited clear limitations. For example, rigid and affine methods (ICP, SICP, CPD_Rigid, CPD_Affine) produced large AE (ranging from 56.7° to 116.8°) and MGE often exceeding 30 mm, indicating substantial orientation and anatomical mismatches. Non-rigid approaches such as CluReg and FFD partially alleviated these errors but still showed unsatisfactory stability, with MGE as high as 68.6 mm and GCE exceeding 80 mm.

Among all tested methods, BCPD++ achieved the most favorable performance: it reduced the MPE to 1.6 mm, the lowest across all methods, and achieved the smallest AE (51.1°) compared with over 90° for ICP and 98.5° for SICP. However, even with BCPD++, the MGE (38.6 mm) and GCE (48.7 mm) remained markedly high, underscoring the difficulty of correcting orientation and key anatomical landmarks through fine registration alone.

These findings highlight a critical limitation: without coarse registration to normalize scale and orientation and to constrain anatomical landmarks, fine registration methods—even advanced non-rigid ones like BCPD++—can ensure moderate global geometric alignment but fail to achieve reliable correction in cardiac orientation and interventricular groove matching. In contrast, the proposed coarse-to-fine strategy first eliminates global discrepancies and then applies local refinement, enabling both millimeter-level accuracy in global geometry and robust alignment in important regions.

To more intuitively demonstrate the MGE and GCE, we plotted a visualization of the MGE and GCE, as shown in Figure 13. We also plotted Figure 14 to visualize the AE.

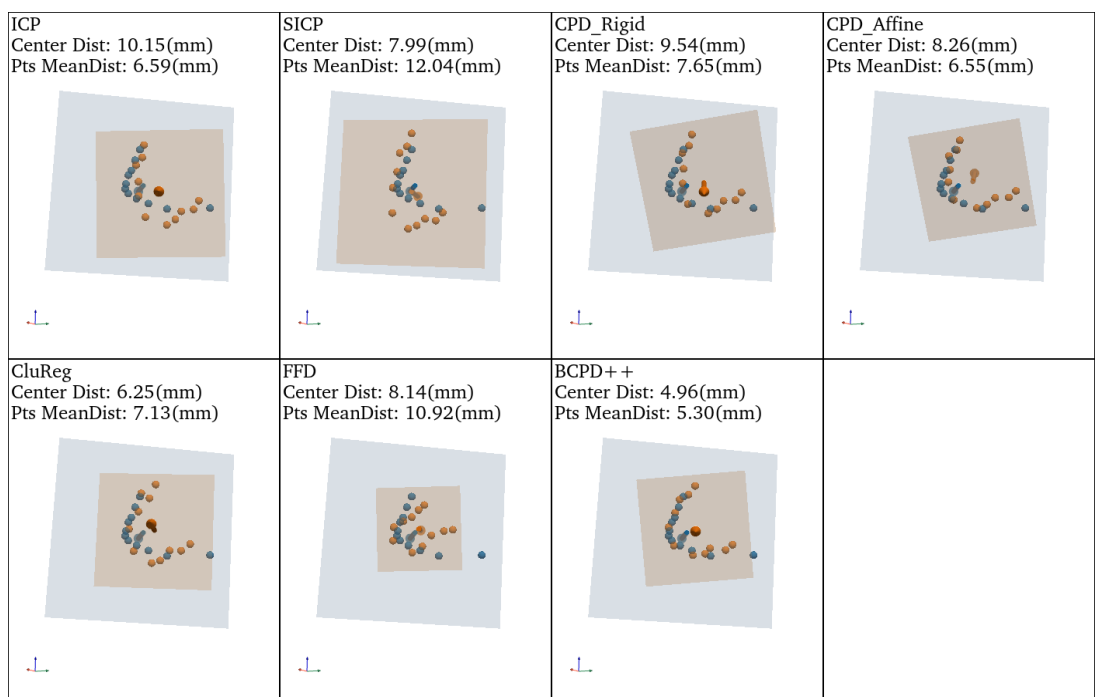

Figure 13. Visualisation of interventricular groove planes for each method, where blue dots represent CTA interventricular groove points, blue arrows represent the direction of CTA interventricular groove planes, orange dots represent SPECT interventricular groove points, and orange arrows represent the direction of SPECT interventricular groove planes.

Figure 13 illustrates the comparative performance of different fine registration methods in the interventricular groove region after coarse alignment. Rigid and affine methods (ICP, SICP, CPD_Rigid) all showed varying degrees of misalignment, with GCE greater than 7 mm and MGE between 6 mm and 12 mm. Notably, SICP produced the least stable results, with the MGE reaching 12.0 mm. Among the non-rigid strategies, BCPD++ achieved the most balanced and robust performance, reducing the MGE to 5.3 mm. This indicates that BCPD++ not only captures local non-rigid deformations effectively but also maintains global stability, yielding anatomically reliable refinement. Although FFD further decreased the GCE to 2.6 mm, its MGE (6.2 mm) suggests less consistent accuracy across the point set. CluReg also showed improvements over rigid methods (GCE 5.9 mm, MGE 6.2 mm), but its performance was slightly inferior to BCPD++.

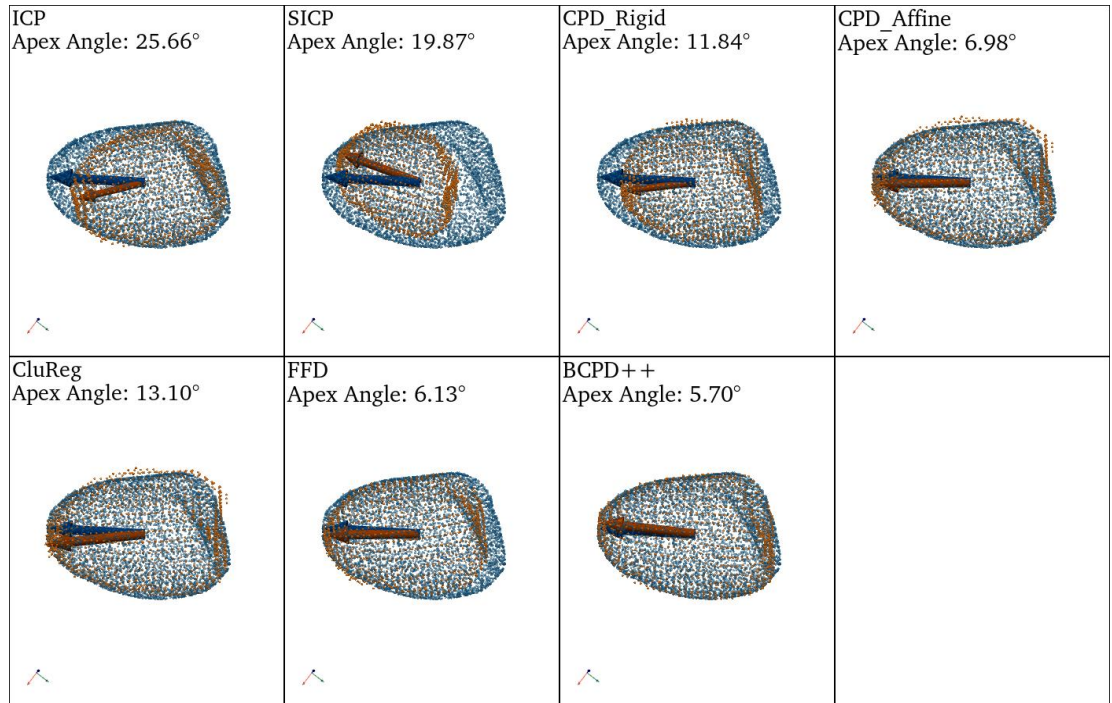

Figure 14. Visualization of the apex displacement angles error for each method, where the blue dots represent the CTA left ventricular outer membrane point cloud, the blue arrows represent the CTA apex direction, the orange dots represent the SPECT left ventricular outer membrane point cloud, and the orange arrows represent the SPECT apex direction.

Figure 14 compares the alignment of apical orientation across methods. SICP and ICP exhibited relatively large deviations, with AE of 19.8° and 25.6°, respectively, suggesting that rigid or scale-rigid methods alone cannot sufficiently correct orientation discrepancies. CPD_Rigid, CPD_Affine and FFD showed improvement, with errors of 11.8°, 6.9° and 6.1°. BCPD++ outperformed all others, further reducing the AE to 5.7°. This demonstrates that BCPD++ not only enhances global geometric consistency but also achieves minimal orientation error in the cardiac long axis, ensuring accuracy when mapping perfusion defects to anatomical segments.

# 4 Conclusion

In this study, we proposed a point cloud-based coarse-to-fine registration approach for high-precision fusion of cardiac SPECT and CTA images. The pipeline couples U-Net segmentation with automated landmark extraction and scale-space normalization for robust cross-modality initialization, then applies multiple fine-registration algorithms separately (ICP, SICP, CPD, CluReg, FFD, BCPD++) for refinement. In a retrospective study of 60 patient cases, the framework achieved millimeter-level accuracy: with BCPD++ attaining the best performance—mean point-cloud distance 1.7 mm, apex displacement angle error 6.7°, mean interventricular-groove distance 8.0 mm, and groove-plane center error 6.6 mm.

Both quantitative and visualization results confirmed that the hierarchical coarse-to-fine strategy effectively overcomes differences in resolution and imaging principles between SPECT and CTA. Moreover, it achieves precise alignment at key anatomical landmarks such as the apex orientation and interventricular groove. The fused images simultaneously present anatomical details from CTA and perfusion information from SPECT within a unified coordinate system.

# Acknowledgments

Fubao Zhu received support from the National Natural Science Foundation of China (Grant Numbers: 62476255, 62303427, 82370513, and 62506343), the Science and Technology Innovation Talent Project of Henan Province University (Grant Number: 25HASTIT028), the Provincial Key Research Project of higher education institutions in Henan (Grant Number: 26A520044), and Zhongyuan Science and Technology Innovation Outstanding Young Talents Program.